\documentclass[review]{elsarticle}
\usepackage{color, soul}
\usepackage[american]{babel}
\journal{Pattern Recognition}

\usepackage{numcompress}
\usepackage{amsmath}
\usepackage{caption}
\usepackage{epstopdf}
\usepackage{subfig}
\usepackage{amsfonts,amssymb}
\usepackage{url}
\usepackage{booktabs} %ÐèÒª¼ÓÔØºê°ü{booktabs}
\usepackage{color}
\usepackage{multirow}
\begin{document}
\captionsetup[figure]{labelfont={bf},name={Fig.},labelsep=period}
\captionsetup[table]{labelfont={bf},name={Tab.},labelsep=period}
\begin{frontmatter}

\title{DLA-Net: Learning Dual Local Attention Features for Semantic Segmentation of Large-Scale Building Facade Point Clouds}
%\tnotetext[mytitlenote]{Fully documented templates are available in the elsarticle package on %\href{http://www.ctan.org/tex-archive/macros/latex/contrib/elsarticle}{CTAN}.}

%% Group authors per affiliation:myfootnote
\author[mymainaddress]{Yanfei~Su}
\author[mymainaddress]{Weiquan~Liu}
\author[mymainaddress]{Zhimin~Yuan}
\author[mymainaddress]{Ming~Cheng\corref{mycorrespondingauthor}}
\ead{chm99@xmu.edu.cn}
\author[mymainaddress]{Zhihong~Zhang}
\author[mymainaddress]{Xuelun~Shen}
\author[mymainaddress]{Cheng~Wang}
%\author[mymainaddress,mysecondaryaddress]{Chenglu~Wen}
%\author[mymainaddress]{Zheng~Zhang}
%\author[mymainaddress]{Congren~Lin}
%\author[mymainaddress,mysecondaryaddress,mythirdaddress]{Jonathan~Li}

\address[mymainaddress]{Fujian Key Laboratory of Sensing and Computing for Smart City, School of Informatics, Xiamen University, Xiamen, China}
%\address[mysecondaryaddress]{Digital Fujian Institute of Urban Traffic Big Data Research, Xiamen University, Xiamen, China}
%\address[mythirdaddress]{Department of Geography and Environmental Management, University of Waterloo, Waterloo, Canada}

%\fntext[myfootnote]{Since 1880.}\fnref{mymainaddress}
%\fntext[myfootnote2]{Since 1880.}
\cortext[mycorrespondingauthor]{Corresponding: Ming~Cheng}

\begin{abstract}
	
Semantic segmentation of building facade is significant in various applications, such as urban building reconstruction and damage assessment. As there is a lack of 3D point clouds datasets related to the fine-grained building facade, we construct the first large-scale building facade point clouds benchmark dataset for semantic segmentation. The existing methods of semantic segmentation cannot fully mine the local neighborhood information of point clouds. Addressing this problem, we propose a learnable attention module that learns Dual Local Attention features, called DLA in this paper. The proposed DLA module consists of two blocks, including the self-attention block and attentive pooling block, which both embed an enhanced position encoding block. The DLA module could be easily embedded into various network architectures for point cloud segmentation, naturally resulting in a new 3D semantic segmentation network with an encoder-decoder architecture, called DLA-Net in this work. Extensive experimental results on our constructed building facade dataset demonstrate that the proposed DLA-Net achieves better performance than the state-of-the-art methods for semantic segmentation.
	
\end{abstract}

\begin{keyword}
	\text{semantic segmentation} \sep{building facade}  \sep {self-attention} \sep {attentive pooling} \sep{DLA-Net}	
\end{keyword}

\end{frontmatter}

%\linenumbers

\section{Introduction}

Automatic semantic segmentation of building facades is extremely important for urban building modeling and such models have a wide range of applications like urban reconstruction~\cite{Sampath2010},\cite{zhang2013layered},\cite{zhang2012tensor},\cite{zhang2013perception} and damage assessment~\cite{Nia2018}.
In the past few years, many works on semantic segmentation of building facades were based on 2D images~\cite{yang2012parsing},\cite{dai2012learning},\cite{martinovic2012three},\cite{tylevcek2013spatial}. However, images acquired by traditional optical imaging-based systems have some intrinsic deficiencies, such as lacking accurate geospatial information, unstable image qualities influenced by illumination conditions, and image distortions caused by camera lens. Furthermore, current semantic segmentation datasets of building facade based on 2D images are insufficient in scale, e.g., the dataset in \cite{tylevcek2013spatial}~only contains about 600 images. Some works used Structure from Motion (SfM) to transform the multi-view 2D images into 3D point clouds before building facades semantic segmentation~\cite{riemenschneider2014learning},\cite{martinovic20153d}. These methods achieve better segmentation quality and higher speed than those based on 2D images. However, the process of converting 2D images to 3D point clouds by the SfM algorithm is time-consuming, and the generated 3D point clouds are sparse in density and limited with the scale of the area.

In recent years, deep learning has achieved great successes in both image interpretation and 3D point cloud processing. However, one of the main bottlenecks of deep learning is that learning a good model requires sufficient manually labeled training data. To our knowledge, there is a lack of large-scale 3D point clouds datasets of fine-grained building facade, which delays the development of deep learning methods in the tasks of 3D point clouds building facade segmentation.

\begin{figure}[!t]
	\centering
	\includegraphics[width=1\linewidth]{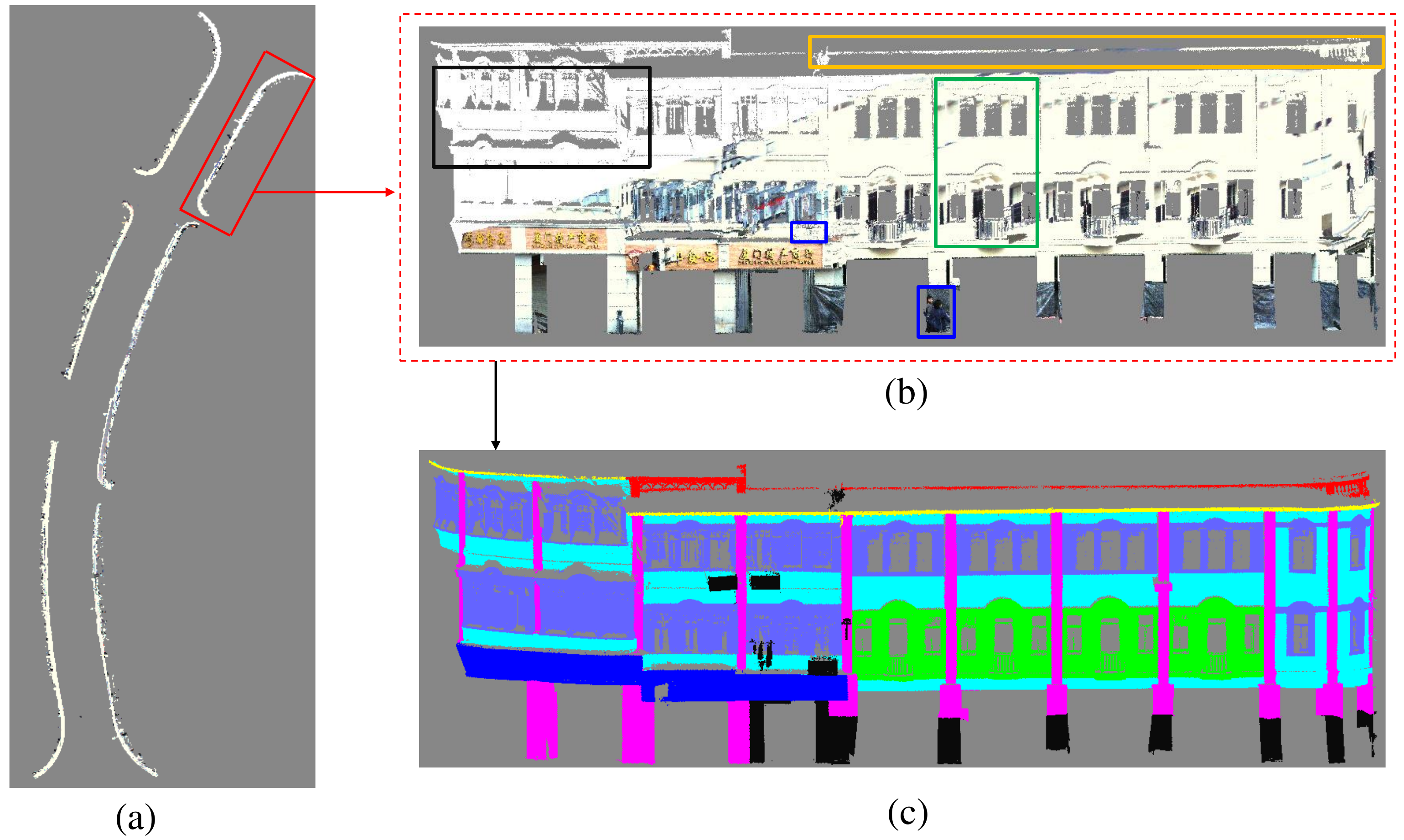}
	\caption{Example of semantic labeling of building facade using color mobile LiDAR point clouds: (a) overhead view of the building facades on both sides of the street; (b) enlarged front view of the building facade within the red box in (a); (c) semantic labeling results.}
	\label{fig:dataset}
\end{figure}

With the rapid development of 3D sensors, large-scale and highly dense 3D point clouds, which provide rich geometric, shape, and texture information, are easily acquired by Mobile Laser Scanners (MLS)~\cite{serna2014paris},\cite{roynard2018paris} in a short time period. By turning to MLS, we aim at closing this data gap to help unleash the full potential of deep learning methods for 3D building facade segmentation tasks. In this paper, we construct the first large-scale fine-grained building facade point clouds dataset benchmark obtained by MLS for semantic segmentation, which covers about three kilometers of city street scenes, and over 160 million manually labeled points. A part of the dataset is visualized in Fig.~\ref{fig:dataset}. We believe that with such a dataset, the research on building facades semantic segmentation would be promoted. Compared with 2D images, 3D point clouds acquired by MLS can provide real-world coordinates of building facades, and immune to the impact of illumination conditions and image distortions. Although 3D point clouds provide rich scene semantic information, as we can see from Fig.~\ref{fig:dataset}(b), segmentation of 3D point clouds is still a challenging task due to the presence of incompleteness (orange bounding box), similar category (green bounding box), color information loss (black bounding box) and occlusion (blue bounding boxes).

In the past few years, a lot of approaches have been proposed for 3D point cloud segmentation, which benefit from the remarkable success of Deep Neural Network (DNN) on irregular point cloud processing. These methods could be roughly divided into three categories~\cite{9127813}: projection-based method~\cite{audebert2016semantic},\cite{lawin2017deep}, \cite{tatarchenko2018tangent},\cite{lin2020fpconv}, discretization-based methods~\cite{long2015fully},\cite{huang2016point},\cite{tchapmi2017segcloud},\cite{dai2018scancomplete}, and point-based methods~\cite{qi2017pointnet},\cite{qi2017pointnet++},\cite{zhao2019pointweb}, \cite{hu2020randla}. Both the projection-based and discretization-based methods are computationally expensive to handle large-scale point clouds, which need extra procedures to transform point clouds to a regular representation and project the intermediate segmentation results back to the point clouds. Due to the generation of intermediates, these methods have not fully exploited the underlying geometric and structural information, which inevitably leads to information loss.

Different from the projection-based and discretization-based methods, point-based methods directly working on 3D point clouds have been proposed. PointNet~\cite{qi2017pointnet}, as the pioneer work, has received much more attention based on which a variety of point-based methods were proposed~\cite{li2018pointcnn},\cite{jiang2018pointsift},\cite{thomas2019kpconv}. Although these methods have achieved promising performances for semantic segmentation, most of them are limited to extremely small 3D point clouds and cannot be directly extended to larger point clouds for two main reasons: (1) These networks usually employ Farthest-Point Sampling (FPS) to obtain keypoint candidates. However, FPS has a time complexity of O($N^{2}$), therefore it is inefficient and time-consuming. (2) Some networks rely on building kernel convolution or graph convolution to learn local features. These methods are also time-consuming and unable to process large-scale point clouds. Lately, RandLA-Net was proposed~\cite{hu2020randla}, which utilized random sampling for candidate selection and subsampling. The time complexity of random sampling is O(1) so that it is highly efficient to process large-scale point clouds. In this paper, we also utilize random sampling for candidate selection and subsampling.

Attention-based methods have been flourishing in recent years~\cite{wang2019exploiting},\cite{zhao2019pooling},\cite{Wang_2019_CVPR},
\cite{feng2020point}. The attention mechanism automatically learns important local features by assigning more weight to the key information. Specially, inspired by the success of transformers in natural language processing~\cite{vaswani2017attention},\cite{wu2019pay} and image analysis~\cite{hu2019local},\cite{ramachandran2019stand}, we develop a method for point clouds processing. The core of the transformer is the self-attention operator, which is invariant to permutation and cardinality~\cite{ramachandran2019stand},\cite{Zhao_2020_CVPR}. So, it is natural to use self-attention as a module in a deep learning network working on 3D point clouds.

In this paper, we aim to design a simple and effective neural network based on an attention module that can directly process large-scale 3D point clouds. The attention module learns Dual Local Attention features from point clouds, which is called DLA module in this paper. The DLA module consists of two blocks, i.e., the self-attention block and the attentive pooling block. First, we aggregate different spatial information to form an enhanced position encoding block, which is embedded in both the self-attention block and the attentive pooling block. Second, the self-attention block concentrates on learning feature representation of the local neighborhoods around each point. Third, the attentive pooling block is used to aggregate local neighboring point feature learned by the self-attention block. The important local features are automatically focused on by the attentive pooling block. Summarily, the proposed DLA module learns more local neighborhood features from point clouds through the self-attention block and the attentive pooling block. Experiments on the building facade dataset show that the network constructed by DLA achieves better segmentation results than the state-of-the-art networks.

In summary, our main contributions in this work include the following:
\begin{itemize}
\item We construct the first large-scale fine-grained building facade 3D point clouds dataset benchmark for semantic segmentation. The dataset will be made public soon\footnote{\url{https://github.com/suyanfei/DLA-Net}} and we believe that the dataset will further boost the research of deep learning on 3D building facade point clouds.
\item We propose an enhanced position encoding block, which aggregates different spatial information to learn more local geometric structure information.
\item We propose the Dual Local Attention (DLA) module, which consists of two blocks including the self-attention block and the attentive pooling block and is able to capture more local neighborhood features of points. The DLA module can be easily applied to various architectures to explore novel point cloud segmentation networks.
\item Extensive experimental results on the building facade dataset demonstrate that the proposed DLA-Net by embedding the DLA module into a standard encoder-decoder architecture achieves state-of-the-art performance.
\end{itemize}

\section{Related Work}
In this section, we introduce the three learning-based categories of large-scale point clouds segmentation methods, including the projection-based, the discretization-based, and the point-based networks~\cite{9127813}.

\subsection{\textbf{Projection-based methods}}
Inspired by the success of 2D Convolutional Neural Networks (CNNs), many existing works~\cite{lawin2017deep},\cite{boulch2017unstructured} project 3D point clouds onto 2D images from multiple virtual views to address the task of semantic segmentation. In a related method, Tatarchenko et al.~\cite{tatarchenko2018tangent}
project the local surface geometry around each point onto a tangent plane, forming tangent images that can be processed by 2D convolution. However, the projection step of these multi-view segmentation methods inevitably introduces information loss of the details and as a result, the underlying geometric and structural information is not fully utilized.

\subsection{\textbf{Discretization-based methods}}
The discretization-based methods usually transform the point clouds into a discrete representation, such as voxels. In~\cite{huang2016point}, the point clouds are divided into voxels and fed to a fully-3D CNN for voxel-wise segmentation. Some methods use the advantage of 3D CNN for point clouds semantic segmentation~\cite{le2018pointgrid},\cite{meng2019vv}. In particular, the Fully-Convolutional Point Network (FCPN)~\cite{rethage2018fully} uses 3D convolutions and weighted average pooling to extract features. The discretization-based methods have more flexibility to process the large-scale point clouds. However, the voxelization step inherently introduces discretization artifacts and information loss.

\subsection{\textbf{Point-based methods}}
 Different from projection-based and discretization-based methods, point-based methods directly work on 3D point clouds. The pioneering work PointNet~\cite{qi2017pointnet} learns a spatial encoding for each point using pointwise MLPs and then aggregates all individual point features as a global representation using symmetrical pooling functions. Based on PointNet, a variety of point-based methods were proposed including~\cite{qi2017pointnet++},\cite{li2018pointcnn},\cite{hu2020randla}, etc. Overall, these methods can be roughly divided into pointwise MLP methods, point convolution methods, RNN-based methods, graph-based methods and attention-based methods.

 \textbf{Pointwise MLP methods}. PointNet++~\cite{qi2017pointnet++}, a hierarchical spatial structure, learns a feature for each point by aggregating the information from local neighboring points. Hu et al.~\cite{hu2020randla} propose an efficient and lightweight network called RandLA-Net for large-scale point cloud segmentation. RandLA-Net utilizes the random point sample method to achieve remarkably high efficiency in terms of memory and computation. Meanwhile, a local feature aggregation module is further proposed to capture and preserve geometric features.

\textbf{Point convolution methods}. A handful of approaches are based on effective convolution operators for point clouds. PointCNN~\cite{li2018pointcnn} transforms neighboring points to the canonical order, which enables traditional convolution to play a normal role. KPConv~\cite{thomas2019kpconv} proposes a spatially deformable point convolution with any number of kernel points which alleviates both varying densities and computational cost, outperforming all associate methods on point clouds segmentation tasks.

\textbf{Graph-based and RNN-based methods}. To capture the underlying shapes and local structure features from point clouds, graph networks and Recurrent Neural Networks (RNN) networks have also been used for semantic segmentation of point clouds. DGCNN~\cite{2018Dynamic} proposes an EdgeConv module, which generates edge features that describe the relationships between a point and its neighbors. RSNet~\cite{2018Recurrent} proposes a lightweight local dependency module and utilizes a slice pooling layer to project the feature of unordered points onto an ordered sequence of feature vectors.

\textbf{Attention-based methods}. In recent years, the powerful attention mechanism has attracted more and more attention~\cite{Wang_2019_CVPR},\cite{feng2020point}. Wang et al.~\cite{wang2019exploiting} first use the graph pointnet module based on graph attention block to dynamically compose and update each point representation within the local point cloud structure, then resorts to the spatial-wise and channel-wise attention strategies to exploit the point cloud global structure. In particular, transformers and self-attention have shown remarkable performance in machine translation and natural language processing~\cite{vaswani2017attention},\cite{wu2019pay}. Inspired by transformer, many works applied a self-attention network into 2D image recognition~\cite{ramachandran2019stand},\cite{hu2019local}. So, in this paper, we try to apply the self-attention network to 3D point clouds of building facades for semantic segmentation.

\section{Method}
In this section, we begin with a brief overview of transformer and self-attention. Then we detail the position encoding block, the self-attention block, and the attentive pooling block for 3D point clouds processing. Lastly, we present our network architecture for 3D semantic segmentation.

\subsection{\textbf{Background}}\label{sec:Bg}
 The deep learning networks based on transformer and self-attention have become a popular and efficient method in the field of natural language processing and 2D image analysis~\cite{vaswani2017attention},\cite{wu2019pay},\cite{hu2019local},\cite{ramachandran2019stand},\cite{Zhao_2020_CVPR}. An attention mechanism uses input-dependent weights to linearly combine the inputs. In~\cite{Zhao_2020_CVPR}, the pairwise self-attention is explored. Mathematically, let $ \chi = \{x_i\}$ be a set of feature vectors, the pairwise self-attention has the following form:
\begin{equation}
	\label{eq:ps}
	  y_i =  \sum\limits_{x_j \in \chi} softmax\bigg(\eta\Big(\delta\big(\alpha(x_i),\beta(x_j)\big)+ \rho\Big)\bigg) \odot \gamma(x_j),
\end{equation}
where $y_i$ is the output feature corresponding to $x_i$, $\odot$ is the Hadamard product, $i$ is the spatial index of feature vector $x_i$. $\alpha$, $\beta$, and $\gamma$ are trainable transformations such as linear projections or MLPs. $\delta$ is a relation function including summation, subtraction, concatenation, Hadamard product, or dot product. The experimental results in~\cite{Zhao_2020_CVPR} show that summation, subtraction, and Hadamard product have the same performance and are better than concatenation and dot product. $\rho$ is a position encoding function and $\eta$ is a mapping function such as MLPs.

\subsection{\textbf{Position encoding block}} \label{sec:PEB}
\begin{figure}[!t]
	\centering
	\includegraphics[width=1\linewidth]{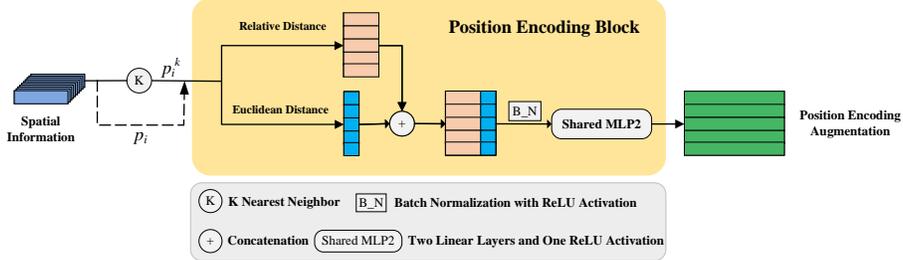}
	\caption{Structure of the position encoding block.}
	\label{fig:position-encoding}
\end{figure}

 Position encoding plays a critical role in networks that are based on transformer and self-attention. In natural language processing, the sine and cosine functions are usually used as position encoding to provide effective position information for sequences~\cite{vaswani2017attention}. In 2D image processing, the relative position of 2D coordinates is used as position encoding to augment image features~\cite{Zhao_2020_CVPR}. In 3D point clouds processing, the 3D point coordinates themselves are a natural candidate for position encoding. We design an enhanced position encoding block based on relative point position, which can better explore the local geometric structure. Given the input clouds with $N$ points, whose $xyz$ positions are denoted as $P={\{p_1},\ldots,{p_i},\ldots,{p_N}\} \subset {\mathbb{R}^{3}}$. For a center point ${p_i}$, its neighboring points, denoted as $\{p{}^1_i,\ldots,p{}^k_i,\ldots,p{}^K_i\}\subseteq P$, are usually gathered by the K-nearest neighbors (KNN) algorithm, which is based on the point-wise Euclidean distances. Our position encoding can be expressed as follows:
\begin{equation}
	\label{eq:pr}
		c{}^k_i  =  MLPs\Big(({p_i} -p{}^k_i) \oplus \parallel{p_i} -p{}^k_i\parallel\Big),
\end{equation}

\noindent where $\oplus$ is the concatenation operation, $\parallel\cdot \parallel$ calculates the Euclidean distance between the neighboring and the center points, and MLPs consist of two linear transformations with a ReLU activation. The visual interpretation of the position encoding block is shown in Fig.~\ref{fig:position-encoding}. Through the position encoding block, the spatial information of point clouds will be enhanced, so that helps the network to learn local features and obtain good performance in practice. Note that the aggregation features need to be processed by batch normalization with ReLU activation operation.

\subsection{\textbf{Self-attention block}} \label{sec:TB}
\begin{figure}[!t]
	\centering
	\includegraphics[width=1\linewidth]{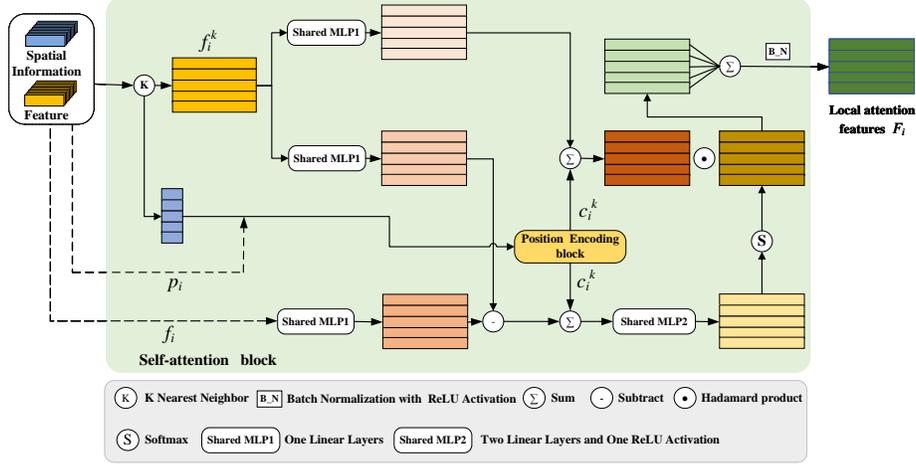}
	\caption{Structure of the self-attention block.}
	\label{fig:transformer-block}
\end{figure}

Self-attention is a natural choice for processing point clouds, which are essentially sets of irregularly embedded metric space. As mentioned in Section ~\ref{sec:Bg}, our self-attention block is based on pairwise self-attention. We choose subtraction as the relation function and add a position encoding to both the mapping function $\eta$ and the transformed features $\gamma$. Our self-attention block is illustrated in Fig.~\ref{fig:transformer-block}. Let the feature of the center point  ${p_i}$ be ${f_i}$, and the feature of its neighboring points $p{}^k_i$ be $f{}^k_i$. ${f_i},f{}^k_i \in  \mathbb{R}^{d}$ ($d$ is feature channels). Our self-attention block is represented as follows:
\begin{equation}
	\label{eq:tb}
	   F_i = \sum\limits_{k=1}^{K} \ softmax\Big(\eta\big(\alpha(f_i)-\beta(f{}^k_i)+c{}^k_i\big)\Big) \odot \big(\gamma(f{}^k_i)+c{}^k_i\big),
\end{equation}

\noindent where $F_i$ is the output feature, $\odot$ is the Hadamard product. $\alpha$, $\beta$, and $\gamma$ are MLP with one linear layer, respectively. The mapping function $\eta$ is an MLP including two linear transformations with a ReLU activation. $c{}^k_i$ is the position encoding , which comes from Eq.~\ref{fig:position-encoding}. The output of the self-attention block $F_i$ is the new set of neighboring features, which explicitly encodes the local geometric structures for the center point $p_i$. The output feature $F_i$ also needs to be processed by batch normalization with ReLU activation operation.

\subsection{\textbf{Attentive pooling block}}

\begin{figure}[!t]
	\centering
	\includegraphics[width=1\linewidth]{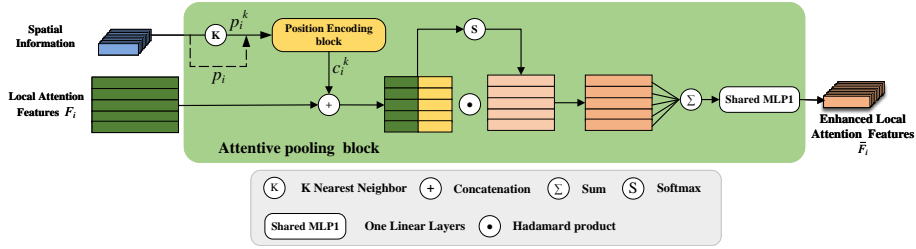}
	\caption{Structure of the attentive pooling block.}
	\label{fig:attentive-pooling}
\end{figure}

It has been proved in recent work~\cite{yang2020robust} that the attention mechanism automatically learns local features. We continually turn to the attention mechanism to further explore the local features. Our structure of the attentive pooling block is shown in Fig.~\ref{fig:attentive-pooling}.

Given the output of the self-attention block $F_i$ and position encoding $c{}^k_i$, we use a relation function to aggregate them.
It is formally defined as follows:
\begin{equation}
	\label{eq:a}
	\hat{F_i} = \eta(F_i, c{}^k_i).
\end{equation}
In the attentive pooling block, we use the concatenation $\oplus$ as the relation function $\eta$.

To get an attention weight for the aggregation feature $\hat{F_i} = \{\hat{f{}^1_i},\ldots,\hat{f{}^k_i},\ldots,\hat{f{}^K_i}\}$, a shared function $\varphi(.,.)$ is designed, which consists of a shared MLP followed by softmax. The learnable weight $W_i$ in the shared MLP assigns a unique attention score to each feature. The learned attention scores can be regarded as a soft mask that automatically selects the important features. Then these features are weighted and summed to generate a new informative feature vector $\bar{F_i}$. This process is represented as follows:
\begin{equation}
	\label{eq:b}
	\bar{F_i} = \sum \limits_{k=1}^{K}  \ \varphi(\hat{f{}^k_i},W_i) \odot \hat{f{}^k_i},
\end{equation}
where $\odot$ is the Hadamard product.

\subsection{\textbf{Architecture of DLA-net}}
\begin{figure}[!t]
	\centering
	\includegraphics[width=1\linewidth]{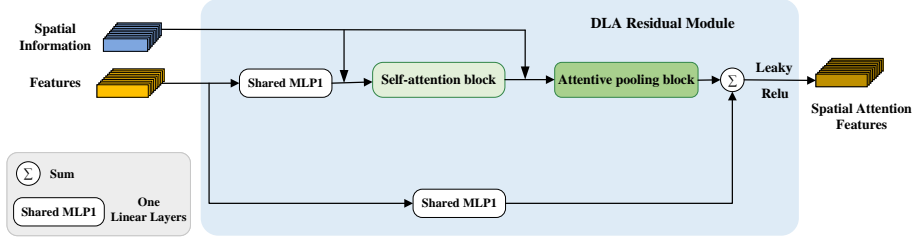}
	\caption{Architecture of the DLA.}
	\label{fig:DLA}	
\end{figure}

The architecture of the proposed DLA module is shown in Fig.~\ref{fig:DLA}. The inputs of DLA include the spatial information and the features learned previously. The spatial information is utilized to construct the point encoding block which is embedded in the self-attention block and the attentive pooling block. The features learned previously are used by the DLA module for exploring the local information. Inspired by the successful ResNet~\cite{he2016deep}, we stack self-attention block and attentive pooling block with a skip connection as a residual module. The DLA residual module is simple but effective and achieves state-of-the-art performance on our constructed large-scale building facade dataset.

\begin{figure}[!t]
	\centering
	\includegraphics[width=1\linewidth]{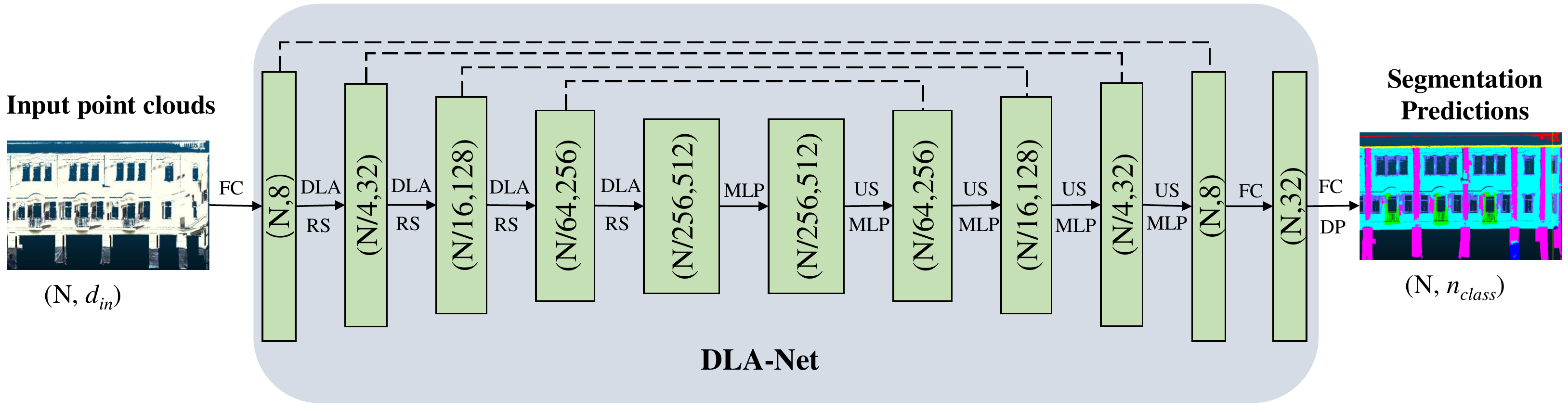}
	\caption{Architecture of the DLA-Net, (N,$d_{in}$) represents the number of points and feature dimension respectively of the input point clouds; FC: Fully Connected layer; DLA: Dual Local Attention residual module; RS: Random Sampling, MLP: shared Multi-Layer Perceptron; US: Up-sampling; (N,$n_{class}$) represents the number of points and classes respectively of the output point clouds.}
	\label{fig:Architecture}
\end{figure}

We embed the proposed DLA residual module into a standard encoder-decoder architecture, resulting in the new segmentation network, DLA-Net. The complete architecture of DLA-net is shown in  Fig.~\ref{fig:Architecture}. The input of the DLA-Net is a large-scale point cloud of size $ N \times d_{in}$ where $N$ is the number of points and $d_{in}$ is the input feature dimension. Each point is presented by its 3D coordinates and color information, i.e., $d_{in}=6$. First, each point is extracted by a fully connected layer, the dimension is unified to 8 as input to the network. The network consists of four encoder layers and four decoder layers. The encoder layers are used to progressive encode the features in which the DLA module and random sampling operation are embedded. The number of points is gradually decreased from $N$ to $ \dfrac{N}{256}$, while the feature dimension is increased from 8 to 512. Second, four decoder layers are used to decode the feature. The decoded features are upsampled through a nearest-neighbor interpolation, and further concatenated with the intermediate feature map produced by encoder layers through skip connections. At last, two shared fully connected layers and a drop layer with ratio 0.5 are used to predict the semantic labels. The outputs of DLA-Net are the predicted semantic labels of all points, with a size of $N \times n_{class}$, where $n_{class}$ is the number of classes.
\section{Experiments}
\subsection{\textbf{Implementation details}}
Our experiments are implemented in Tensorflow on a server with an Intel (R) Xeon (R) E5-2683 CPU, 64GB of RAM, an NVIDIA Titan X GPU, CUDA10.0, and cuDNN v7. We use the Adam optimizer with an initial learning rate of $10^{-2}$. The DLA-Net is trained for 100 epochs, with the learning rate dropped by $5\%$ after each epoch. The cross-entropy loss is used for training. The batch size is set as 6. The number of nearest points $K$ is set as 16. A fixed number of points (40960) are sampled from each point cloud and fed into the network while training. The whole raw point cloud is input during testing.

\subsection{\textbf{Dataset}}
\begin{figure}[!t]
	\centering \large
	\includegraphics[width=1\linewidth]{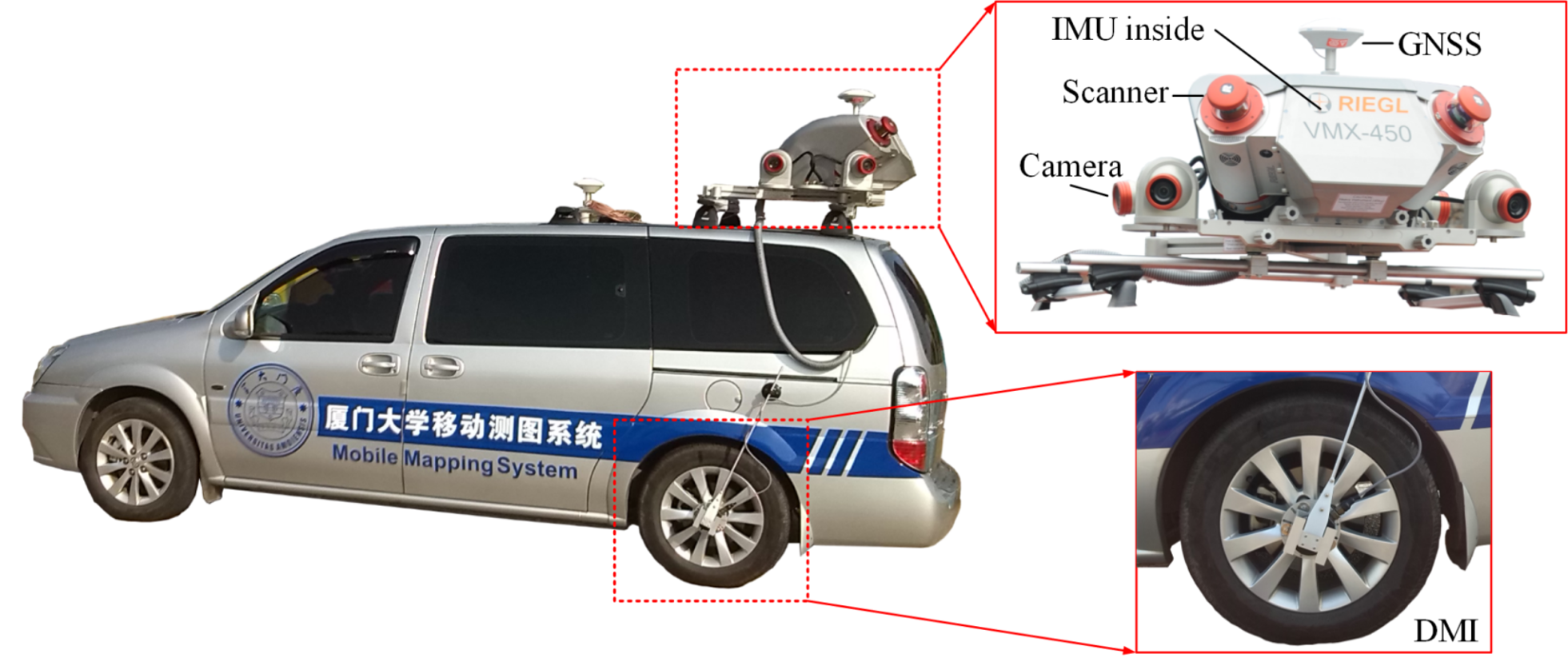}
	\caption{Illustration of RIEGL VMX-450 mobile LiDAR system and its configurations.}
	\label{fig:450}
\end{figure}

In this paper, the fine-grained building facade point clouds dataset was acquired along the National Road and Siming South Road in Xiamen, China, by a RIEGL VMX-450 mobile LiDAR system. The RIEGL VMX-450 system~\cite{450} smoothly integrates two RIEGL VQ-450 laser scanners, a global navigation satellite system (GNSS) antenna, an inertial measurement unit (IMU), a distance measurement indicator (DMI), and four high-resolution digital cameras (see Fig.~\ref{fig:450}). This integrated set was mounted on the roof of a minivan with an average speed of 40-50 km/h. After data acquisition, RiProcess, a post-process software released by RIEGL corporation, is used to calibrate the images with point clouds for the generation of colorized building facade point clouds. The scanning frequency of the VMX-450 scanning system is up to 400Hz, the pulse frequency is 1100kHz, the maximum scanning distance is up to 800m, the scanning accuracy is up to 5 mm, and the positioning accuracy is up to 5 cm. Therefore, the data obtained by the RIEGL VMX-450 onboard mobile laser scanning system is high-resolution 3D laser point clouds data.

\begin{figure}[!t]
	\centering \large
	\includegraphics[width=0.8\linewidth]{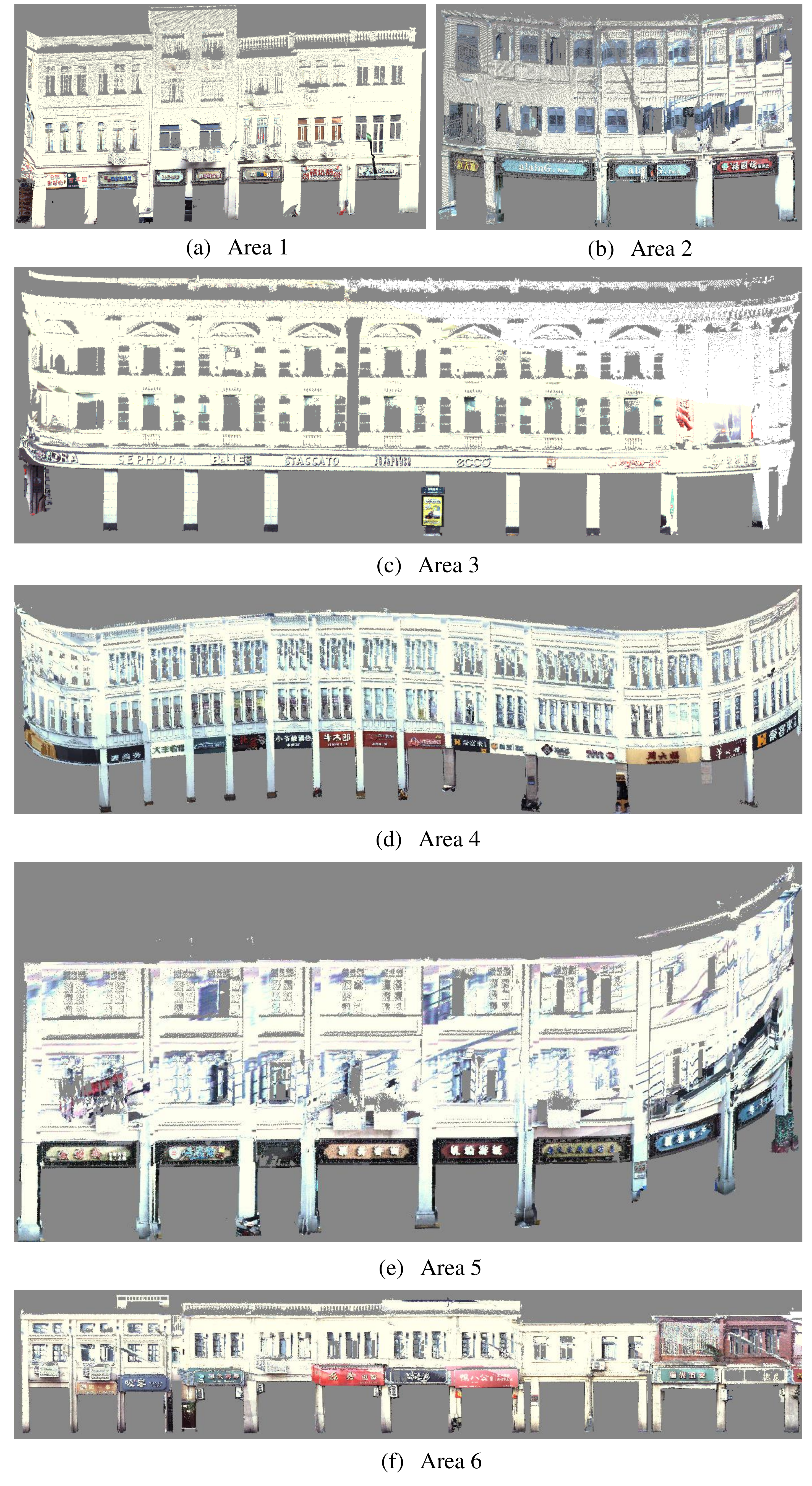}
	\caption{Illustration of a representative building facade for each area.}
	\label{fig:bf}
\end{figure}

\begin{figure}[!htbp]
	\centering \large
	\includegraphics[width=0.85\linewidth]{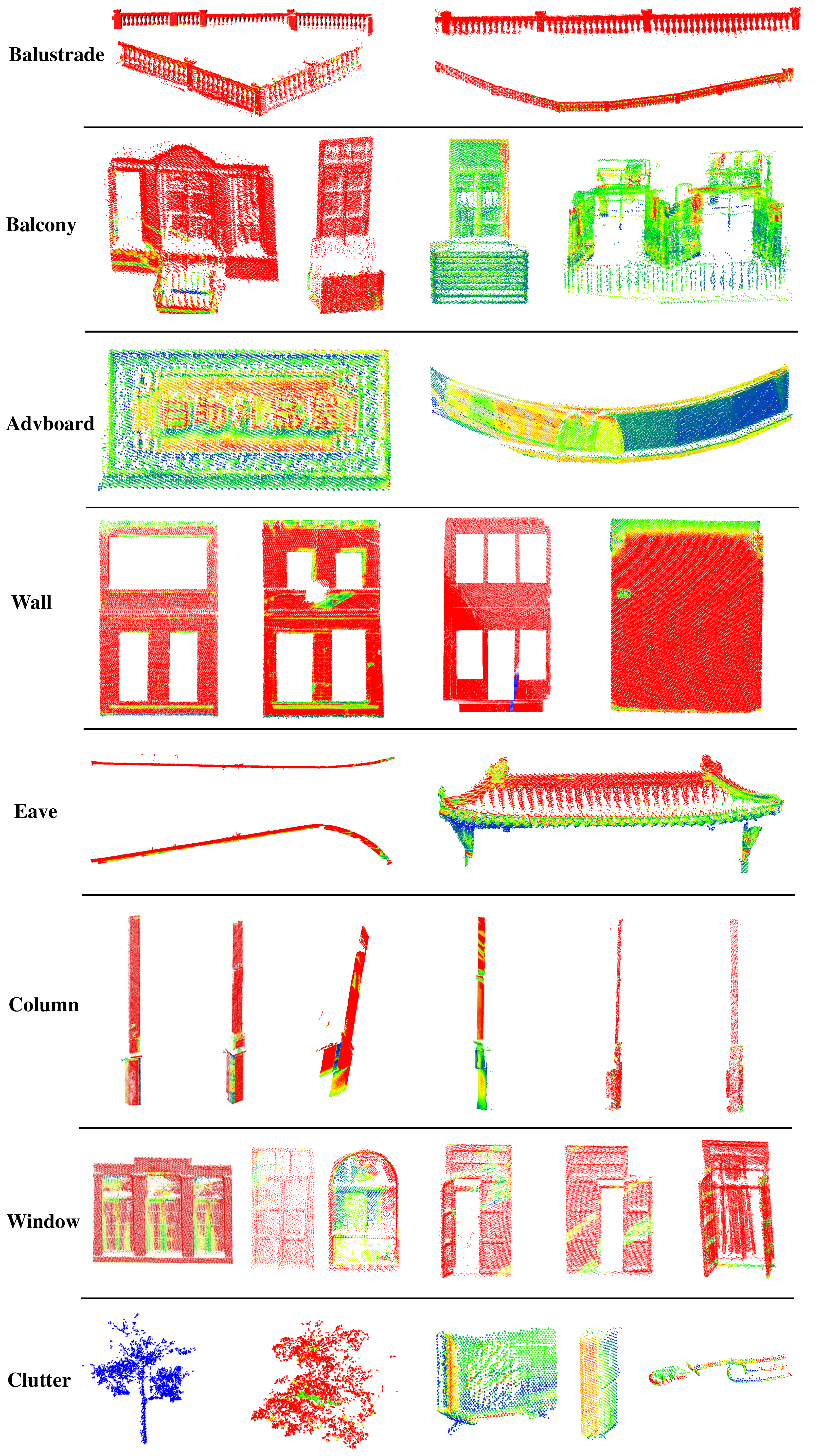}
	\caption{Illustration of each category from different perspectives and with different forms.}
	\label{fig:visual}
\end{figure}

The colorized building facade point clouds dataset has a point density of about 7,000 points/m$^2$ and covers two roads sections of total about 3,000 meters. In a real city street scene, the point clouds acquired by the RIEGL VMX-450 system have many non-building facade points such as ground, pedestrians, vehicles, trees, outlier, and so forth. So it is necessary to preprocess the data by manually removing non-building facade point clouds. The existence of these points inevitably interferes with the collection of building facade points and causes incomplete data. After data preprocessing, we divide the dataset into 6 areas according to the style and location of the building facade. We use the CloudCompare tool to visualize a representative building facade for each area (see Fig.~\ref{fig:bf}). As seen from Fig.~\ref{fig:bf}, the building facade in the dataset is the typical commercial street scene. Tab.~\ref{tab:area number} details the number of points in each area and the total of six areas.

\begin{table}[!htbp] \centering \scriptsize
	\setlength{\tabcolsep}{2.3mm}{	
		\caption{The number of points in each area and the total of six areas.}
		\label{tab:area number}
		\begin{tabular}{l|cccccc}
			\toprule
			& Area 1 & Area 2 & Area 3 & Area 4 & Area 5 & Area 6\\
			\midrule
			Points   &33,206,410 & 30,063,042 & 14,790,151 &24,954,180 & 22,785,852 &32,959,555  \\
			\midrule
			Total &\multicolumn{6}{c}{\textbf{158,759,190}}  \\
			\bottomrule
	\end{tabular}}
\end{table}

\begin{table}[!htbp] \centering \scriptsize
	\setlength{\tabcolsep}{0.8mm}{
		\caption{The number of points in each category.}
		\label{tab:category number}
		\begin{tabular}{l|cccccccc}
			\toprule
			& Balustrade & Balcony & Advboard & Wall & Eave & Column & Window & Clutter\\
			\midrule
			Points  & 1,535,374 & 7,169,298 & 22,190,094 &51,651,937 & 5,744,710 &40,034,103 & 26,287,509 & 4,146,165 \\
			\midrule
			Total &\multicolumn{8}{c}{\textbf{158,759,190}}  \\
			\bottomrule
	\end{tabular}}
\end{table}

 we construct the ground truth for the dataset by manually and thoroughly classifying all points into the following eight categories: balustrade, balcony, advboard (i.e., advertising board), wall, eave, column, window, and clutter. Tab.~\ref{tab:category number} details the number of points in each category. The statistical data show the challenge of category imbalance. In order to better intuitively understand each category, we selectively visualize each category from different perspectives and with different forms in Fig.~\ref{fig:visual}. Specially, we regard convex parts and windows as balconies when they are vertically distributed in space. We regard the air conditioner external unit hanging on the wall,  weeds and bushes growing on the eaves, wire boxes and street lamps hanging on the column as clutter. The samples of each category are diverse in shape, which shows the richness of our building facade dataset.

Challenges, such as object occlusions, incompleteness, category imbalance, and diversity of category samples, commonly exist in the building facade dataset. These challenges make the semantic segmentation of building facade point clouds a difficult task.

\subsection{\textbf{Evaluation on the dataset}}
To fully evaluate our DLA-Net on building facade dataset for semantic segmentation, we used two modes: (a) Area 1 is used for test and Areas 2 to 6 are used for train, similarly, each area is used for test in turn and others are used for train. (b) k-fold cross-validation (k=6). We use overall pointwise accuracy (OA), mean classwise intersection over union (mIoU), and mean of classwise accuracy (mAcc) as the standard metrics.
In addition, 8 methods including PointNet~\cite{qi2017pointnet}, PointNet++~\cite{qi2017pointnet++}, DGCNN~\cite{2018Dynamic}, ELGS~\cite{wang2019exploiting}, RSNet~\cite{2018Recurrent}, PointCNN~\cite{li2018pointcnn}, KPConv~\cite{thomas2019kpconv}, RandLA-Net~\cite{hu2020randla} are used for the referred approaches which are retrained on our dataset.
\begin{table}[!htbp] \centering \scriptsize
	\setlength{\tabcolsep}{0.6mm}{
		\caption{Quantitative results of different approach on the building facade dataset (Area 1, Area 2, and Area 3 take turns as the test ).}
		\label{tab:1-3 in turn-validation}
		\begin{tabular}{l|ccc|ccc|ccc}
			\toprule
			\multirow{2}{*}{methods} & \multicolumn{3}{|c|}{Area 1--test} & \multicolumn{3}{|c|}{Area 2--test} & \multicolumn{3}{c}{Area 3--test} \\ \cline{2-10}
			& OA($\%$) & mIoU($\%$) & mAcc($\%$)  & OA($\%$) & mIoU($\%$) & mAcc($\%$) & OA($\%$) & mIoU($\%$) & mAcc($\%$)\\
			
			\midrule
			PointNet~\cite{qi2017pointnet} &55.5 & 35.0 &	51.7 &	62.3 &	41.8 &	57.6 &	58.7 &	36.2 &	54.3
			 \\
			PointNet++~\cite{qi2017pointnet++} & 58.6 &	38.5 &	56.3 &	60.7 &	34.4 &	53.3 &	62.9 &	39.4 &	57.8
			\\
			DGCNN~\cite{2018Dynamic} & 63.1 &	42.1 &	57.7 &	69.4 &	51.0 &	66.7 &	66.5 &	45.2 &	63.1
			 \\
			ELGS~\cite{wang2019exploiting} &69.1 & 51.5 &	62.1 &	68.9 &	47.0 &	58.3 &	65.5 &	39.3 &	55.6
			\\
			RSNet~\cite{2018Recurrent} & 68.7 &	50.0 &	65.4 &	69.0 &	50.2 &	64.3 &	70.0 &	47.6 &	65.3
			\\
			PointCNN~\cite{li2018pointcnn} & 77.0 & 55.9 & 68.8 & 79.6 &	52.0 &	65.7 &	79.1 &	53.4 &	68.3
			\\
			KPConv~\cite{thomas2019kpconv} &\textbf{85.7}  &\textbf{68.3} &	78.3 &\textbf{84.7}	 &	66.8 &	77.8 &\textbf{83.5} &56.9 &	67.1
			\\
			RandLA-net~\cite{hu2020randla} &	80.0 &	63.0 &	77.4 &	81.6 &	62.0 &	77.2 &	80.0 &	55.8 &\textbf{73.1}	
			\\	
			\midrule
			\textbf{DLA-Net(ours)} & 83.9 &	68.2 & \textbf{81.4} & 84.3 &\textbf{66.9}  & \textbf{81.5} & 83.1 & \textbf{58.4} &	71.9  \\
						
			\bottomrule
	\end{tabular}}
\end{table}

\begin{table}[!htbp] \centering \scriptsize
	\setlength{\tabcolsep}{0.6mm}{
		\caption{Quantitative results of different approach on the building facade dataset (Area 4, Area 5, and Area 6 take turns as the test ).}
		\label{tab:4-6 in turn-validation}
		\begin{tabular}{l|ccc|ccc|ccc}
			\toprule
			\multirow{2}{*}{methods} & \multicolumn{3}{|c|}{Area 4--test} & \multicolumn{3}{|c|}{Area 5--test} & \multicolumn{3}{c}{Area 6--test} \\ \cline{2-10}
			& OA($\%$) & mIoU($\%$) & mAcc($\%$)  & OA($\%$) & mIoU($\%$) & mAcc($\%$) & OA($\%$) & mIoU($\%$) & mAcc($\%$)\\
			
			\midrule
			PointNet~\cite{qi2017pointnet} 	& 51.3 	&27.3 	&39.4 	&54.3 	&30.8 	&45.5 	&47.3 	&24.7 	&36.2 	\\
			
			PointNet++~\cite{qi2017pointnet++}  & 44.2 	&21.8 	&33.4 	&63.4 	&39.4 	&54.5 	&57.2 	&32.3 	&45.0 \\
			
			DGCNN~\cite{2018Dynamic} & 60.0 	& 35.1 	& 49.4 	& 66.4 	& 41.0 	& 55.5 	& 55.4 	& 29.7 	& 42.3 	\\
			ELGS~\cite{wang2019exploiting} &52.8 	&28.8 	&45.2 	&69.8 	&51.6 	&63.2 	&62.3 	&36.6 	&49.2  \\
			RSNet~\cite{2018Recurrent}  & 56.5 &	34.5 &	50.4 &	69.1 	&43.5 &	58.0 &	60.5 	&36.2 	&52.8 \\
			PointCNN~\cite{li2018pointcnn} & 70.1 	&41.3 	&53.2 	&78.3 	&50.7 	&64.4 	&65.6 	&38.4 	&52.1 \\
			KPConv~\cite{thomas2019kpconv} &77.1 &51.2 	&62.7 &	77.9 &	52.4 &	66.8 &\textbf{78.4} &	\textbf{51.7} 	& \textbf{63.0 }	\\
			RandLA-net~\cite{hu2020randla} &\textbf{79.4} 	&53.9 &	\textbf{69.4} &	79.8 &	60.3 &	71.0 &	72.1 	& 47.7 &	62.2 \\	
			\midrule
			\textbf{DLA-Net(ours)} &78.5 &\textbf{54.3} &	69.1 &\textbf{81.2} &\textbf{63.6 }	&\textbf{74.0}	&	76.0 & 50.4 & 62.2   \\
			
			\bottomrule
	\end{tabular}}
\end{table}
The quantitative results of all the referred methods tested on each area are reported in Tab.~\ref{tab:1-3 in turn-validation} and Tab.~\ref{tab:4-6 in turn-validation}. As seen from Table.~\ref{tab:1-3 in turn-validation}, when Area 1 is used for the test, our method performs better than others on mAcc, and only 0.1$\%$ less than KPConv on mIoU. Our DLA-Net achieves better performance than state-of-the-art methods on mIoU when the test set are Area 2 and Area 3.
As seen from Tab.~\ref{tab:4-6 in turn-validation}, when the test set is Area 4, our method has the best mIoU among all the methods. DLA-Net outperforms all prior models according to all metrics when Area 5 is used for the test. In general, DLA-Net achieves better performance on at least one metric except for Area 6.
\begin{table}[!htbp] \centering \tiny
	\setlength{\tabcolsep}{0.8mm}{
		\caption{Semantic segmentation results on the our data set, evaluated with 6-fold cross-validation.}
		\label{tab:cross-validation}
		\begin{tabular}{l|ccc|cccccccc}
			\toprule
			methods & OA($\%$) & mIoU($\%$) & mAcc($\%$) & Abalustrade & balcony & advboard & wall & eave & column & window & clutter\\
			\midrule
			PointNet~\cite{qi2017pointnet} & 53.3 &	32.8 &46.9 & 12.2 &	15.6 & 45.3 & 31.8 & 39.4 & 56.1 & 30.7 & 31.4 \\
			PointNet++~\cite{qi2017pointnet++} & 57.8 &	34.3 &	50.0 &	39.4 &	12.5 &	41.5 &	46.0 &	45.9 &	44.8 &	36.7 &	8.5 \\
			DGCNN~\cite{2018Dynamic}	&61.8 	&40.2 	&54.4 &	14.8 &	22.9 &	51.0 &	42.2 &	55.0 &	60.6 &	40.5 &\textbf{34.3} \\
			ELGS~\cite{wang2019exploiting}	&63.0 &	40.6 &	54.8 &	18.2 &	21.3 &	56.4 &	46.4 &	51.4 &	58.0 &	43.4 &	29.2 \\
			RSNet~\cite{2018Recurrent}	 & 63.9 & 42.1 & 56.0 	&19.7 &	21.8 &	62.7 &	44.4 &	55.6 &	63.2 &	43.4 &	25.7 \\
			PointCNN~\cite{li2018pointcnn} 	&75.0 &	48.6 &	62.1 &	43.6 &	9.9 &	70.5 &	60.3 &	58.6 &	76.9 &	53.4 &	15.6 \\
			KPConv~\cite{thomas2019kpconv} & \textbf{81.3} &	58.0 &	68.0 &	49.7 &	35.7 &	77.5 &\textbf{67.9}	 &	\textbf{58.7} &\textbf{83.7} & 60.4 & 30.4 \\
			RandLA-net~\cite{hu2020randla}	& 78.6 &	56.0 &	68.6 &	51.4 &	36.4 &	78.3 &	62.9 &	58.5 &	76.9 &	58.8 &	24.9 \\	
			\midrule
			\textbf{DLA-Net(ours)} & 81.0 &\textbf	{59.7} & \textbf{71.6}	 &	\textbf{58.2}&\textbf	{43.3} &\textbf{79.1} & 67.5 &	58.0 	&79.0 	&\textbf{62.0} 	&30.2 \\
			
			\bottomrule
	\end{tabular}}
\end{table}

Then, we report the quantitative results of all the referred methods evaluated with 6-fold cross-validation in Tab.~\ref{tab:cross-validation}. As seen from Tab.~\ref{tab:cross-validation}, DLA-Net has the best mIoU and mAcc among all these methods and outperforms the prior state-of-the-art by nearly 3$\%$ in mIoU and 4.4$\%$ in mAcc. As for OA, it is slightly lower than KPConv, but it is better than all the other compared methods. DLA-Net also achieves the best performance on 4 categories, including balustrade, balcony, advboard, and window.

In the Section~\ref{sec:PEB} and~\ref{sec:TB}. The batch normalization with ReLU activation plays an important part in the position encoding block and the self-attention block. We define $\times$  as without batch normalization and ReLU activation, and $\surd$ as batch normalization with ReLU activation. The experiments are conducted on the building facade dataset, tested on  Area 1.

\begin{table}[!htbp] \centering \normalsize
	\setlength{\tabcolsep}{1mm}{
		\caption{The results of DLA-net by  whether to use batch normalization with ReLU activation in the Position Encoding Block and  self-attention block.}
		\label{tab:ablation-bn}
		\begin{tabular}{cc|ccc}
			\toprule
			Position encoding block & Self-attention block  & OA(\%) & mIoU(\%)  & mAcc(\%)  \\
			\midrule
			$\times$ & $\times$ & 82.2 & 64.0 & 76.0 \\
			$\times$ & $\surd$ & 83.4 & 66.2 &78.0 \\
			$\surd$ & $\times$ & 81.5 & 64.8 & 78.5 \\
			$\surd$ & $\surd$ & \textbf{83.9} & \textbf{68.2} & \textbf{81.4} \\
			\bottomrule
	\end{tabular}}
\end{table}

As we can see from Tab.~\ref{tab:ablation-bn}, when batch normalization with ReLU activation operation is no used in both the Position Encoding Block and self-attention Block, DLA-Net achieves poor performance. When batch normalization with ReLU activation operation is used in the Position Encoding Block or self-attention Block, DLA-Net performs better than not used. Only When batch normalization with ReLU activation operation is used in both the Position Encoding Block and self-attention Block, DLA-Net outperforms all prior models.

The presence of color information is helpful to improve the accuracy of semantic segmentation. Tab.~\ref{tab:rgb} presents the quantitative results of DLA-Net with respect to the different types of input point clouds. The experiments are conducted on the building facade dataset, evaluated with 6-fold cross-validation. When DLA-net is trained given both point coordinates and
RGB information, the network achieves significantly better segmentation accuracy.
\begin{table}[!htbp] \centering \tiny
	\setlength{\tabcolsep}{0.8mm}{
		\caption{Quantitative results of DLA-Net on our building facade dateset.}
		\label{tab:rgb}
		\begin{tabular}{l|ccc|cccccccc}
			\toprule
			methods & OA($\%$) & mIoU($\%$) & mAcc($\%$) & balustrade & balcony & advboard & wall & eave & column & window & clutter\\
			\midrule
			DLA-Net(w/o RGB) & 80.0 &	57.6 &69.4 & 55.1 &	39.3& 76.8 & 66.2 & 56.4 & 78.9 & 60.4 & 27.4 \\
			
			DLA-Net(w RGB) & \textbf{81.0} &\textbf	{59.7} & \textbf{71.6}	 &	\textbf{58.2}&\textbf{43.3} &\textbf{79.1} &\textbf{67.5} &	\textbf{58.0}  & \textbf{79.0}	&\textbf{62.0} 	&\textbf{30.2} \\
			
			\bottomrule
	\end{tabular}}
\end{table}

\begin{figure}[!t]
	\centering \large
	\includegraphics[width=0.9\linewidth]{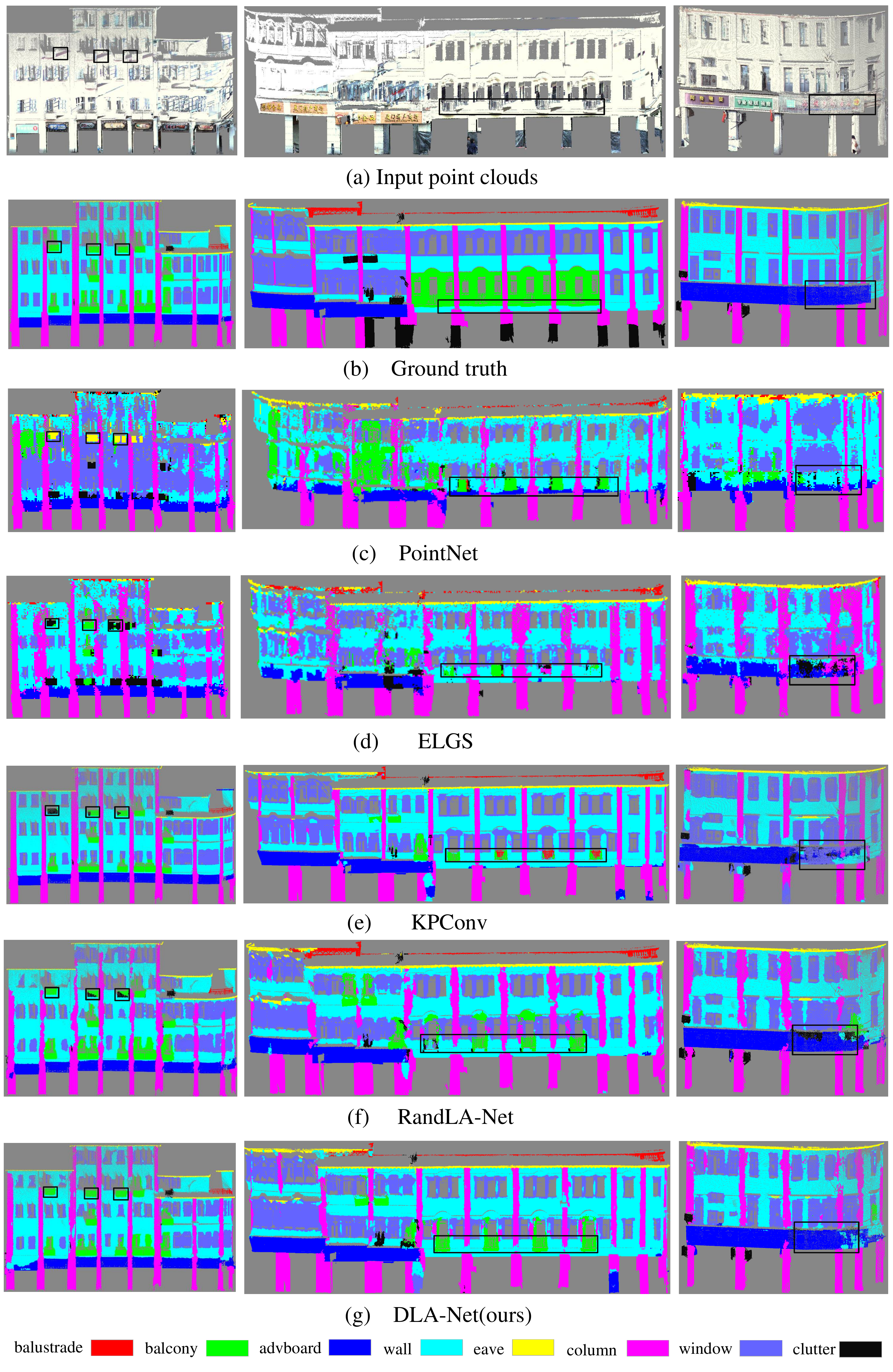}
	\caption{Qualitative comparison on the three scenes of building facade dataset. Different colors denote different categories. These scenes contain 8 categories. Black boxes highlight some examples where our method performs better than others.}
	\label{fig:segmentation}
\end{figure}

Lastly, the segmentation results visualization examples of several typical buildings facade are shown in Fig.~\ref{fig:segmentation}.
The figure qualitatively compares the semantic segmentation obtained by PointNet~\cite{qi2017pointnet}, ELGS~\cite{wang2019exploiting}, KPConv~\cite{thomas2019kpconv}, RandLA-Net~\cite{hu2020randla}, and our DLA-Net.
We can see that objects such as the clutter hung on the wall, balcony and window embedded in the wall, clutter cover on the column are quite difficult to segment.
As can be seen from Fig.~\ref{fig:segmentation}, it is obvious that the segmentation results of PointNet and ELGS are the worst.
 We use black bounding boxes to highlight some examples where our method performed significantly better than the competitors. In the first building facade, the convex part of the balcony is very similar to the air conditioner external unit in the spatial shape but different in size making the semantic segmentation a real challenge. So, PointNet~\cite{qi2017pointnet}, ELGS~\cite{wang2019exploiting}, KPConv~\cite{thomas2019kpconv}, and RandLA-Net~\cite{hu2020randla} mistake the convex part of the balcony as clutter. The proposed DLA-Net can identify the convex part of the balcony more accurately than these networks. In the second building facade, all the networks failed to separate the clutter from the column, because the clutter is covered on the column. PointNet~\cite{qi2017pointnet}, ELGS~\cite{wang2019exploiting}, KPConv~\cite{thomas2019kpconv}, and RandLA-Net~\cite{hu2020randla} also did not fully identify the convex part of the balcony. This is because the convex part of the balcony is diverse in shape, which is also similar to the balustrade. However, the proposed DLA-Net can still accurately identify the convex part of the balcony. In the third building facade, the output of DLA-Net on windows is more regular than that of PointNet~\cite{qi2017pointnet}, ELGS~\cite{wang2019exploiting}, KPConv~\cite{thomas2019kpconv}, and RandLA-Net~\cite{hu2020randla}, and DLA-Net performs better on advboard than others.

\subsection{\textbf{Ablation study}}
The effectiveness of DLA-Net is verified by the experimental results on the building facade dataset. To better understand the DLA-Net, we further evaluate it and conduct the following three groups of experiments. we also use OA, mIoU and mAcc as the metrics.

\subsubsection{\textbf{Ablation study on position encoding block}}

The following ablation studies are conducted to study the impacts of the position encoding block in our framework. Position encoding block is a combination of different spatial information. So, we explore the effects of position encoding composed of different spatial information in our DLA-Net. The experiments are conducted on the building facade dataset, evaluated with 6-fold cross-validation. There are six combinations as follows:

(1) Position encoding block consists of the neighboring points $ p{}^k_i$ only.

(2) Position encoding block consists of the relative position $p_i - p{}^k_i $ only.

(3) Position encoding block consists of the relative position  $p_i - p{}^k_i $ and Euclidean distance $\parallel{p_i} - p{}^k_i\parallel$.

(4) Position encoding block consists of  the  points  $ p_i$, the relative position $p_i - p{}^k_i $, and Euclidean distance $\parallel{p_i} - p{}^k_i\parallel$.

(5) Position encoding block consists of  the neighboring points $ p{}^k_i$, the relative position $p_i - p{}^k_i $ , and Euclidean distance $\parallel{p_i} - p{}^k_i\parallel$.

(6) Position encoding block consists of  the  points  $ p_i$, the neighboring points $ p{}^k_i$, the relative position $p_i - p{}^k_i $ , and Euclidean distance $\parallel{p_i} - p{}^k_i\parallel$.
\begin{table}[!htbp] \centering \normalsize
	\setlength{\tabcolsep}{0.8mm}{
		\caption{The results of DLA-net by concatenating the different spatial information as the position encoding block.}
		\label{tab:ablation-position encoding}
		\begin{tabular}{l|ccc}
			\toprule
			Position encoding block & OA(\%) & mIoU(\%)  & mAcc(\%)  \\
			\midrule
			$ p{}^k_i$ & 69.5 & 47.0 & 61.9 \\
			$p_i - p{}^k_i $  &80.8 &59.3& 71.1\\
			$(p_i - p{}^k_i) \  \oplus \parallel{p_i} - p{}^k_i\parallel$(\textbf{ours})& \textbf{81.0}&\textbf{59.7} & \textbf{71.6} \\
			$p_i \oplus  (p_i - p{}^k_i) \ \oplus \parallel{p_i} - p{}^k_i\parallel$ & 80.9&58.9&70.0 \\
			$p{}^k_i \oplus  (p_i - p{}^k_i) \  \oplus \parallel{p_i} - p{}^k_i\parallel$ &80.6 & 59.2 & 71.2 \\
			$ p_i \oplus p{}^k_i \oplus \ (p_i - p{}^k_i) \  \oplus \parallel{p_i} - p{}^k_i\parallel$ &80.4 &58.5 & 69.9 \\
			\bottomrule
	\end{tabular}}
\end{table}

The results are shown in Tab.~\ref{tab:ablation-position encoding}. We can see that: 1) Only using the neighboring points $ p{}^k_i$ as the position encoding block is not going to work very well. 2) The combinations based on the relative point position used for the position encoding block get good performance. As we can see from this, the relative position plays an important role in the position encoding block. 3) When the relative position and Euclidean distance are concatenated as the position encoding block, our DLA-Net achieves the best performance. 4) Base on the relative position and Euclidean distance, if the points $p_i$, or the neighboring points $p{}^k_i$, or together, is concatenated as the position encoding block, the segmentation result is worse than 3). It is because too much spatial information interferes with each other.

\subsubsection{\textbf{Ablation study on self-attention block}}
In section~\ref{sec:TB}, the position encoding block is added to the mapping function and the transformed features. In this section, we investigate the location of the position encoding block in the self-attention block. The experiments are conducted on the building facade dataset, tested on Area 1.

\begin{table}[!htbp] \centering \normalsize
	\setlength{\tabcolsep}{0.8mm}{
		\caption{Ablation study on self-attention pooling block.}
		\label{tab:ablation-self-attention}
		\begin{tabular}{l|ccc}
			\toprule
			Self-attention block  & OA(\%) & mIoU(\%)  & mAcc(\%)  \\
			\midrule
			None & 83.3 & 66.1 & 77.5 \\
			Add only to  the mapping function  block & 83.5 & 67.6 &80.5 \\
			Add only to  transformed features  & 83.3 & 66.4 & 79.2 \\
			\textbf{Ours} & \textbf{83.9} & \textbf{68.2} & \textbf{81.4} \\
			\bottomrule
	\end{tabular}}
\end{table}
The results are shown in Tab.~\ref{tab:ablation-self-attention}. It can be viewed that without position encoding block in the self-attention block, the performance of DLA-Net drops. When the position encoding block is added only to the mapping function or only to the transformed features, the performance of DLA-Net is also not good. Only when the position encoding block is both added to the mapping function and the transformed features, the self-attention block perform well.

\subsubsection{\textbf{Ablation study on attentive pooling block}}
We verify the effectiveness of attentive pooling block from many aspects, including removing all of it, only removing the position encoding block of it, replace it with max-pooling or avg-pooling. The experiments are conducted on the building facade dataset, tested on Area 1.

\begin{table}[!htbp] \centering \normalsize
	\setlength{\tabcolsep}{0.8mm}{
		\caption{Ablation study on attentive pooling block.}
			\label{tab:ablation-attentive pooling}
			\begin{tabular}{l|ccc}
				\toprule
				Attentive pooling block  & OA(\%) & mIoU(\%)  & mAcc(\%)  \\
				\midrule
				Remove all & 83.1 & 64.7 & 76.2 \\
				Remove position encoding block & 82.9 & 65.1 &77.2 \\
				Replace with max-pooling & 79.2 & 62.1 & 77.9 \\
				Replace with avg-pooling & 77.8 & 61.0 & 76.7 \\
				\textbf{Ours} & \textbf{83.9} & \textbf{68.2} & \textbf{81.4} \\
				\bottomrule
		\end{tabular}}
	\end{table}
As we can see from Tab.~\ref{tab:ablation-attentive pooling}, the results prove two points: 1) the attentive pooling block plays an important role in DLA; 2) the position encoding block plays an important role in the attentive pooling block.
\section{Conclusion}
In this paper, we construct the first large-scale fine-grained building facade point clouds dataset, which can be used to facilitate research on visual tasks related to building facade. In order to fully explore the local neighborhood features in the 3D point cloud on semantic segmentation task, we present a novel attention-based network DLA-net in which the self-attention block and attentive pooling block based on the powerful attention mechanism are used for learning important local features. DLA module could be easily embedded into various network architectures for point cloud segmentation and we embed it into an encoder-decoder architecture, resulting in the DLA-Net in this work. Extensive experiments on the building facade dataset benchmarks demonstrate the state-of-the-art performance of our proposed DLA-Net. In the future, the dataset can be used not only for the task of semantic segmentation, but also for many computer vision tasks such as building facade point clouds completion and 3D reconstruction.

%\section*{References}

\bibliography{mybibfile}

\end{document}